\documentclass[letterpaper]{article} 
\usepackage{aaai25}  
\usepackage{times}  
\usepackage{helvet}  
\usepackage{courier}  
\usepackage[hyphens]{url}  
\usepackage{graphicx} 
\usepackage{natbib}  
\usepackage{caption} 
\usepackage{amsmath}
\usepackage{amssymb}
\usepackage{algorithmic}
\usepackage{multirow}
\usepackage{algorithm}
\usepackage{amsfonts}
\usepackage{scrextend}
\usepackage{amsopn}
\usepackage{adjustbox}
\usepackage[T1]{fontenc}
\usepackage[utf8]{inputenc}

\usepackage{graphicx}
\usepackage{physics}
\usepackage{amssymb}

\usepackage{microtype}
\usepackage{inconsolata}
\nocopyright

\usepackage{newfloat}
\usepackage{listings}
\DeclareCaptionStyle{ruled}{labelfont=normalfont,labelsep=colon,strut=off} 
\floatstyle{ruled}
\newfloat{listing}{tb}{lst}{}
\floatname{listing}{Listing}
\title{Controlling Equational Reasoning in Large Language Models \\ with Prompt Interventions}
\author{
    Jordan Meadows\textsuperscript{\rm 1}, 
    Marco Valentino\textsuperscript{\rm 2}, 
    Andr\'e Freitas\textsuperscript{\rm 1, 2, 3}
}

\affiliations{
    \textsuperscript{\rm 1}Department of Computer Science, University of Manchester\\
    \textsuperscript{\rm 2}Idiap Research Institute\\
    \textsuperscript{\rm 3}National Biomarker Centre, CRUK-MI\\
    jordan.meadows@postgrad.manchester.ac.uk\\
    \{marco.valentino, andre.freitas\}@idiap.ch\\
}

\usepackage{bibentry}

\begin{document}

\maketitle

\begin{abstract}
This paper investigates how hallucination rates in Large Language Models (LLMs) may be controlled via a symbolic data generation framework, exploring a fundamental relationship between the rate of certain mathematical errors and types of input intervention. Specifically, we systematically generate data for a derivation generation task using a symbolic engine, applying targeted interventions to prompts to perturb features of mathematical derivations such as the surface forms of symbols, equational tree structures, and mathematical context. We then evaluate the effect of prompt interventions across a range of LLMs including fine-tuned T5 models, GPT, and LLaMa-based models. Our experiments suggest that T5-Large can outperform the few-shot performance of GPT-4 on various evaluation sets generated via the framework. However, an extensive evaluation based on human analysis, template-based error detection, and text generation metrics reveals model weaknesses beyond what the reference-based metrics singularly describe. We use these results to tie characteristic distributional footprints of interventions to the human evaluation of LLM derivation quality, potentially leading to significant control over fine-grained mathematical capabilities of language models with respect to specific types of errors.

\end{abstract}

%

\section{Introduction}

\begin{figure*}[h!]
    \centering
    \includegraphics[width=0.9\textwidth]{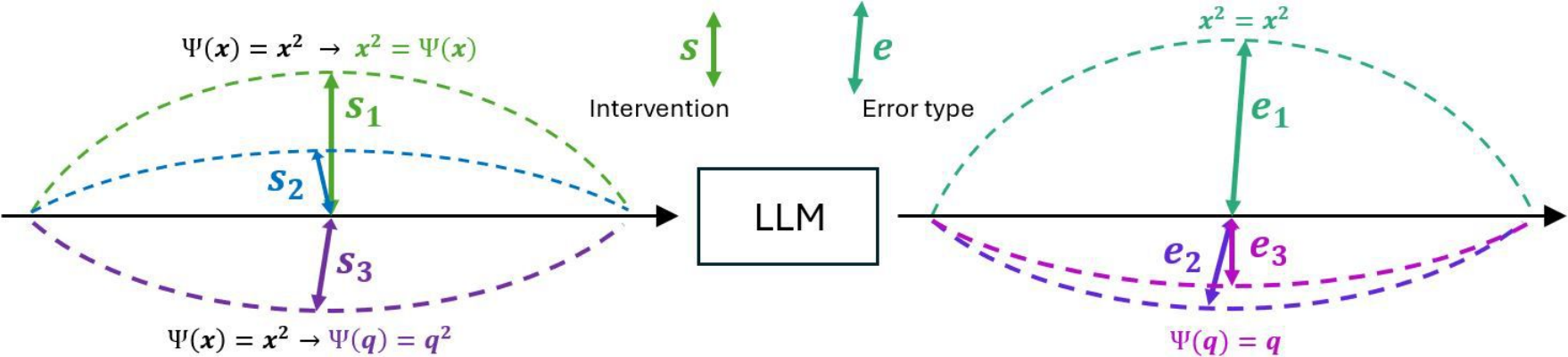}
    \caption{Mechanism relating the type and strength of a set of interventions to their corresponding mathematical errors. The black line indicates the original prompt/output, while the arrows and dashed lines indicate the extent of the input perturbation and respective output delta.}
    \label{fig:overall}
\end{figure*} 

Large Language Models (LLMs) possess the potential to accelerate mathematical discovery~\cite{trinh2024solving}. Yet, without incorporating symbolic approaches, their ability to derive correct mathematical results is significantly impeded by their sensitivity to input perturbations and tendency to hallucinate~\cite{mirzadeh2024gsm,frieder2023mathematical}. In order to understand and mitigate persisting challenges in mathematical reasoning, recent work has focused on the ability of LLMs to learn and sequentially apply symbolic operations to premise equations~\cite{chen2024premise,valentino2024multioperationalmathematicalderivationslatent}, investigating whether state-of-the-art models can derive goal equations defined within prompts~\cite{meadows2024symbolicframeworkevaluatingmathematical}. Such \textit{derivation-style} equational reasoning is at the core of many applied mathematical fields, such as theoretical physics, engineering, and quantitative finance~\cite{plaisted1993equational,premtoon2020semantic}. 

In this paper, we aim to establish a deeper connection between input perturbations applied to prompts and certain mathematical hallucinations, attempting to expose a fundamental relationship between training data, prompt intervention type, and derivational error distributions in LLM output. In particular, we are interested in whether such a relationship can be identified and potentially leveraged to control the rate of certain types of mathematical errors.

To this end, we develop a methodology supported by a symbolic data generation framework that has been effectively applied in related contexts~\cite{valentino2024multioperationalmathematicalderivationslatent,meadows2024symbolicframeworkevaluatingmathematical}, which we adapt and improve ($\approx 15$x faster) to construct and augment various \textit{fine-grained} datasets for equational reasoning tasks. 
This framework allows us to operate at a mathematical granularity within derivations that is far greater than what is surfaced on published derivations~\cite{mann2018manipulating, akrobotu2022qubo}, where many steps are typically omitted or summarised. This ultimately addresses a fundamental incompleteness problem for reasoning data available for training and evaluating LLMs~\cite{villalobos2022run}. Given that fine-grained workings contribute to much of the theoretical research distilled in papers, that generative models frequently hallucinate when solving domain-specific reasoning problems~\cite{shuster2021retrieval,taylor2022galactica,frieder2023mathematical,wysocka2023large,meadows2024exploring}, and that granular reasoning lends itself better to explainability and inference control~\cite{hebenstreit2023automatically,yao2023tree,yuan2023well,valentino2024nature}, it is clear that methods of control over the rate of mathematical reasoning errors made by LLMs, without incorporating external solvers~\cite{toshniwal2024openmathinstruct,liu2024augmenting,trinh2024solving,jiang2022draft,quan-etal-2024-verification}, are highly valuable. 

By leveraging the granularity of the symbolic data generation framework, we investigate control mechanisms that involve fine-tuning LLMs and applying targeted prompt interventions that systematically manipulate inputs to the models. These interventions --- here, alterations in symbolic representation, equational structure, and contextual elements —-- serve as levers to induce and regulate specific error types in model outputs, moving beyond methodologies which are agnostic to certain classes of mathematical hallucination and/or deal with less complex equation manipulation~\cite{stolfo2023causalframeworkquantifyrobustness,meadows2024symbolicframeworkevaluatingmathematical}. Crucially, our experiments suggest that certain interventions distinctly correspond to distributional footprints in the error space, opening up the possibility to better investigate the underlying nature of mathematical hallucinations in language models.

Overall, our contribution can be summarised as follows:\footnote{Code, data, hyperparameters, and further experimental details available at: \url{https://github.com/jmeadows17/deriving-equations-with-LLMs}}

\begin{enumerate}
    \item We construct and release a dataset of 30k mathematically fine-grained prompt-derivation pairs spanning 18 operators, 155 wildcard (LaTeX) symbols, 4 targeted distribution shifts, up to 10 equations per derivation, and 160k steps --- all developed using a symbolic data generation framework. We also improve the speed of this framework by $\approx 15$x and fix limitations involving irrelevant steps.

    \item We demonstrate that T5 models fine-tuned on fine-grained mathematical derivations generated using our approach can match or surpass the in-distribution few-shot performance of GPT-4 according to all evaluation methods. In addition, we evaluate a range of open source models in a few-shot setting including LLaMa-2-7B and Llemma-7B.

    \item We investigate 3 \textit{complementary evaluation methods} to determine the mathematical proficiency of LLMs: \textbf{(1.)} reference-based text generation metrics (including 4 metrics), \textbf{(2.)} template-based detection of mathematical errors, and \textbf{(3.)} human evaluation of 750 derivations. Each method is applied to both in-distribution and out-of-distribution data augmented by interventions.
    While we find that template-based and human evaluations are correlated, they both \textit{strongly disagree} with reference-based metrics, proving that the latter deliver misleading model performance rankings and inappropriate representations of the relative effect of interventions.

    \item We demonstrate a fundamental underlying mechanism where the rates of certain errors are controlled by systematically varying both the strength and type of interventions on the prompt, visualised in Fig.~\ref{fig:overall}. For instance, the rate of ``redundant'' equations (\textit{e.g.,} $x = x$) increases by up to 2000\% in fine-tuned models, based on an intervention which perturbs equation symmetry. An intervention that removes integration/differentiation results from the prompt leads to a relative increase in step skipping by up to 300\%, and leads GPT-4 to make 1000\% more reasoning errors, according to human evaluation. Although each intervention naturally affects multiple error categories, their distributional footprint can be uniquely identified, and controlled via the magnitude of each intervention type.
\end{enumerate}

\section{Derivation Generation with LLMs}
\label{derivation_generation}

Given a goal equation $G$ and premises $\mathcal{P}$, that are arranged within some prompt template $t(\mathcal{P}, G)$, we aim to assess the ability of an LLM to systematically apply a set of symbolic operations to premises to generate a sequence of equations $\mathcal{\hat{D}}$, which represents a reasonable derivation of $G$. Given an LLM $\mathcal{M}$, a derivation is generated through $\mathcal{M}:t(\mathcal{P}, G) \mapsto \mathcal{\hat{D}}$. An idealised metric $M^*(\mathcal{D}^*,\hat{\mathcal{D}})$ scores a derivation, where $\mathcal{D}^*$ is ideally a valid human written derivation corresponding to input prompt $t(\mathcal{P}, G)$. We generally aim to optimise
\begin{equation*}
    \mathcal{M}^* = \underset{\mathcal{M}}{\mathrm{argmax}}; M^*\big(\mathcal{D}^*, \mathcal{M}(t(\mathcal{P}, G))\big).
\end{equation*}

 \noindent However, we do not have access to ideal derivations $\mathcal{D}^*$ corresponding to templates $t(\mathcal{P}, G)$, nor ideal metric $M^*$ suitable for scoring $\hat{\mathcal{D}}$. Instead, we employ a symbolic engine to \textit{approximate} ground truth derivations to obtain $\tilde{\mathcal{D}^*}$ (Alg.~\ref{alg:derivation_generation}). Moreover, we evaluate over a sample of derivations. This means that, in practice, we are instead finding $\mathcal{M}^*$ such that
\begin{equation*}
\scalebox{0.85}{
    $\mathcal{M}^* = \underset{\mathcal{M}}{\mathrm{argmax}}; \frac{1}{N}\sum_{i=1}^N M\big(\tilde{\mathcal{D}^*}_i, \mathcal{M}(t(\mathcal{P}_i, G_i))\big),$
}
\end{equation*}

\noindent where $N$ is the sample size. In this work, we consider $M$ as a reference-based generation metric to automatically evaluate derivations, but we contrast this with a human evaluation based on equation consistency and coherent operator usage, and a template-based error detection method. 

\section{Dataset Construction}

The symbolic data generation process used in this work is described by Alg.~\ref{alg:derivation_generation}.
Given a vocabulary of symbols $\mathcal{V}$ and a set of computer algebra operations $\mathcal{R}$, our goal is to generate a mathematical derivation $\mathcal{D}$ represented by an ordered list of steps $s_i \in \mathcal{D}$. In order to construct $\mathcal{D}$, an initial reasoning step $s_1 = (\text{premise}, \text{annotation})$ is generated such that $\mathcal{D} = [s_1]$. Subsequently, an operation $r \in \mathcal{R}$ is sampled, which in its most general form accepts two operands (arity 2). The first operand is an equation $s_{j,1}$ from tuple $s_j \in \mathcal{D}$. A suitable secondary variable ($\in \mathcal{V}$), expression, or equation operand $m$ is extracted from $\mathcal{D}$, and the next equation is generated by applying operation $r$ through $s_{i+1, 1} = r(s_{j,1}, m)$. The annotation $s_{i+1,2}$ is a list containing the name of the operation, the equation index, and the secondary operand, such that $s_{i+1,2} =[r, j, m']$ (where $m'$ is a variable/expression string or equation index representing operand $m$). Therefore, step $s_{i+1} = (r(s_{j,1}, m), [r, j, m'])$. If $\mathcal{D} = [s_1]$, then $i = j = 1$, and the derivation updates such that $\mathcal{D} = [s_1, s_2]$. This process repeats until the derivation reaches a target length.

\begin{algorithm}[t]
\caption{Derivation Generation}
\label{alg:derivation_generation}
\textbf{Input}: Vocabulary of symbols $\mathcal{V}$, Set of operations $\mathcal{R}$\\
\textbf{Output}: Ordered list of derivation steps $\mathcal{D}$

\begin{algorithmic}[1] 
\STATE Initialize derivation $\mathcal{D}$ with a premise step $s_1 = (\text{premise equation}, \text{annotation})$
\STATE Set $i = 1$
\WHILE{desired length of $\mathcal{D}$ not reached}
    \STATE Sample an operation $r \in \mathcal{R}$
    \STATE Select an equation $s_{j,1}$ from tuple $s_j \in \mathcal{D}$
    \STATE Extract a suitable operand $m$ from $\mathcal{V}$ or $\mathcal{D}$ that matches the requirements of $r$
    \STATE Generate the next equation $s_{i+1, 1} = r(s_{j,1}, m)$
    \STATE Create an annotation $s_{i+1,2}$ representing the operation and operands: $s_{i+1,2} = [r, j, m']$ where $m'$ is an index or variable/expression string corresponding to $m$
    \STATE Append the new step to the derivation: $\mathcal{D}.\text{append}((s_{i+1,1}, s_{i+1,2}))$
    \STATE Increment $i$
\ENDWHILE
\STATE \textbf{return} $\mathcal{D}$
\end{algorithmic}
\end{algorithm}

{\renewcommand{\arraystretch}{1.1}%
\begin{table}[t]
	\centering
	\scalebox{1}{
		\begin{tabular}{@{}c|c @{}}
			\textbf{Dataset} & \textbf{Size} (k)\\
			\hline
            Training & 15.3 \\
            Static Test Set (In-distribution) & 3.1 \\
            Variable Renaming (VR) & 2.9 \\
            Expression Exchange (EE) & 3.1 \\
            Alternative Goal (AG) & 3.1 \\
            Step Removal (SR) & 1.0\\
		\end{tabular}
		}
	\caption{Sizes for the various Derivation Generation datasets.}
	\label{tab:sizes}
\end{table}}

\begin{figure}[t]
    \centering
    \includegraphics[width=0.8\linewidth]{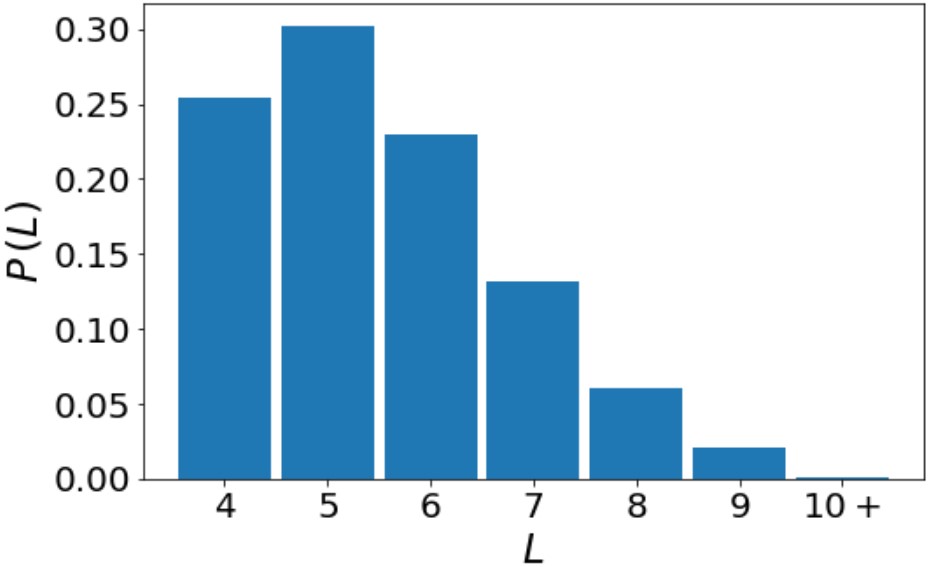}
    \caption{Length distribution $P(L)$ of derivations.}
    \label{fig:length_distribution}
\end{figure}

{\renewcommand{\arraystretch}{1.5}%
\begin{table}[t]
	\centering
	\scalebox{0.8}{
		\begin{tabular}{@{}c|c|c|c @{}}
			\textbf{Length $(L)$} & \textbf{Permutations} & \textbf{Chain} & \textbf{P(Chain)}\\
			\hline
            \multirow{2}{*}{4} & \multirow{2}{*}{842} & $\partial \rightarrow \partial_E \rightarrow S_L$ & 0.0369\\
              &     & $\int \rightarrow \int_E \rightarrow S_L$ & 0.0186\\
              \hline
            \multirow{2}{*}{5} & \multirow{2}{*}{2850} & $+ \rightarrow \partial \rightarrow \partial_E \rightarrow S_L$ & 0.0053\\
              &      & $- \rightarrow \partial \rightarrow \partial_E \rightarrow S_L$ & 0.0048
		\end{tabular}
		}
	\caption{For a given derivation length $L$, \textbf{Permutations} describes the number of unique operation sequences present in the training data. \textbf{Chain} describes the two most frequent operation sequences based on symbols. \textbf{P(Chain)} is the probability of the chain.}
	\label{tab:chains}
\end{table}}

\noindent A derivation generated from Alg.~\ref{alg:derivation_generation} is then \textit{perturbed} according to a set of interventions to form out-of-distribution examples. Specifically, given the task of Derivation Generation instantiated via the prompt template $t = t(\mathcal{P}, G)$ and a ground truth derivation $\mathcal{D}$, a static dataset $X$ consisting of $(t, \mathcal{D})$ is constructed using the process described above. Subsequently, a perturbed dataset $X_n$ is formed by applying a perturbation function $P_n$ to all $(t, \mathcal{D}) \in X$ to form $(t', \mathcal{D}') \in X_n$, such that $P_n: X \rightarrow X_n$, and $n$ denotes the number of perturbations considered. 

Tab.~\ref{tab:sizes} describes the dataset sizes generated by the symbolic framework, Fig.~\ref{fig:length_distribution} displays the distribution of equation counts in ground truth references (\textit{i.e.,} derivation length), and Tab.~\ref{tab:chains} shows that the operation chains responsible for forming the underlying derivation reasoning do not frequently repeat.

\paragraph{On the omission of natural language.} Although the framework (through Alg.~\ref{alg:derivation_generation}) outputs \textit{both} equations and step annotations by default, we purposefully remove annotations from the output in the specific Derivation Generation task considered in this work. Firstly, annotations give additional information on the dependency structure between equations, but they are certainly not necessary for the purpose of generating valid dependency graphs between equations with generative models. Ground truth derivations from the dataset can be clearly followed without natural language. Second, without annotations, the coherence of the derivation depends on the equations and their dependencies alone. This is more targeted than the alternative of additionally determining whether annotations match up with equations.

\paragraph{Improvements to symbolic data generation framework.} We rely on the symbolic framework proposed in \citet{meadows2024symbolicframeworkevaluatingmathematical} to support the experimental pipeline following related work in non-generative settings~\cite{valentino2024multioperationalmathematicalderivationslatent}. We improve the data generation approach in the following ways:

\begin{enumerate}
    \item \textbf{Support for complex LaTeX symbols} (\textit{e.g.,} $\Psi_{nl}$) instead of more basic symbols (\textit{e.g.,} $x$).  

    \item \textbf{Removed irrelevant and disconnected equations} from derivations by including additional dependency checks between derivation steps. This improvement was crucial for eliciting the desired derivational behaviour in models via fine-tuning and in-context learning. 

    \item \textbf{Improved runtime efficiency by a factor of 15} by allowing derivations to both equal or exceed the target length, including timeout decorators on certain operations, and using more efficient iteration limiting. The approximate difference is \textbf{< 0.05} min/derivation compared to \textbf{0.7} min/derivation tested over 100 samples. 
\end{enumerate}

\noindent We adopt the same set of hyperparameters described in \citet{meadows2024symbolicframeworkevaluatingmathematical} using the following values: p\_history=10, p\_arity\_0=5, p\_renaming=1, p\_arity\_1=50, p\_evaluate=50, p\_arity\_2=100, p\_int\_or\_diff=1, p\_subs=5.

\subsection{Prompt Interventions and Perturbations}

A perturbation or intervention is a transformation applied to the input text and/or ground truth that ideally changes a single target textual aspect. We apply 4 interventions to the static test set to generate corresponding perturbed sets.
\paragraph{Variable Renaming (VR).} In the training set, derivations rely on a vocabulary of 155 symbols (\textit{e.g.,} $\Psi_{nl}$, $E_n$, $\mathbf{J}_P$, $\eta$, $g^{\prime}_{\varepsilon}$). For each example in the static set, we uniquely map each symbol to an out-of-distribution symbol sampled from 11 Greek letters (\textit{e.g.,} $E_n = n + x$ becomes $\alpha = \beta + \gamma$).

\paragraph{Expression Exchange (EE).} In the training set and applied mathematics in general, there is an asymmetry with respect to premises being defined with functions on the LHS and expressions on the RHS (\textit{e.g.,} $E_n(n, x) = n + x$). However, operations are frequently used that can substitute LHS for RHS (and vice versa) in many cases, and both functions and operations may appear on either side of equations. Simply, we swap expressions either side of the equality for all equations in the static test set (\textit{e.g.,} $E_n = n + x$ becomes $n + x = E_n$).

\paragraph{Alternative Goal (AG).} For each example in the static set, we derive an alternative goal equation from the penultimate equation, by random selection of operators and operands, which equates to the synthetic data algorithm skipping its first choice goal equation for that derivation. This perturbation should not result in significant differences in model outputs as it simply applies alternative in-distribution operations that occur frequently during training or within few-shot prompts.

\paragraph{Step Removal (SR).} In the training set, equations that occur as a result of evaluating differentials and integrals are included in the prompt as intermediate steps. These are used to guide model outputs. This perturbation removes such \textit{``then derive''} equations from the prompt, which forces models to either circumvent such steps or derive them during inference.

\renewcommand{\arraystretch}{1.2}%
\begin{table*}[t]
    \centering
    \scalebox{0.7}{
        \begin{tabular}{@{}l|c|c|c|c|c|c|c|c|c|c|c|c|c|c|c|c|c|c|c|c@{}}
            & \multicolumn{5}{c|}{ROUGE} & \multicolumn{5}{c|}{BLEU} & \multicolumn{5}{c|}{BLEURT} & \multicolumn{5}{c}{GLEU} \\
            \hline
            & \textbf{S} & \textbf{VR} & \textbf{EE} & \textbf{AG} & \textbf{SR} & \textbf{S} & \textbf{VR} & \textbf{EE} & \textbf{AG} & \textbf{SR} & \textbf{S} & \textbf{VR} & \textbf{EE} & \textbf{AG} & \textbf{SR} & \textbf{S} & \textbf{VR} & \textbf{EE} & \textbf{AG} & \textbf{SR} \\
            \hline
            T5-base $(f)$         & 88.6 & 80.2 & 86.2 & 88.3 & 77.0 & 81.3 & 74.8 & 78.2 & 80.9 & 64.2 & 70.5 & 67.7 & 67.3 & 67.4 & 51.5 & 83.4 & 76.1 & 80.4 & 83.1 & 69.2 \\
            FLAN-T5-base $(f)$    & 87.3 & 24.4 & 84.3 & 86.7 & 77.7 & 79.4 & 41.1 & 76.0 & 78.8 & 66.6 & 68.9 & 18.7 & 67.0 & 67.9 & \textbf{56.8} & 81.7 & 44.2 & 78.5 & 81.3 & 71.0 \\
            T5-large $(f)$        & 89.4 & \textbf{85.0} & 86.8 & 89.2 & 77.7 & 82.8 & \textbf{79.3} & 79.5 & 82.5 & 66.4 & 72.1 & \textbf{70.8} & 68.3 & 69.6 & 54.1 & 84.7 & \textbf{80.8} & 81.5 & 84.4 & 70.6 \\
            FLAN-T5-large $(f)$   & \textbf{90.2} & 83.0 & \textbf{87.1} & \textbf{89.5} & \textbf{78.6} & \textbf{84.6} & 78.5 & \textbf{80.4} & \textbf{83.5} & \textbf{68.9} & \textbf{73.2} & 69.0 & \textbf{68.7} & \textbf{70.3} & 56.1 & \textbf{86.1} & 79.6 & \textbf{82.1} & \textbf{85.1} & \textbf{72.4} \\
            \hline
            T5-base               & 89.5 & 82.2 & 87.3 & 89.9 & 79.9 & 82.8 & 77.2 & 81.6 & 83.7 & 68.8 & 70.5 & 71.1 & 69.6 & 70.1 & 56.5 & 84.4 & 78.0 & 82.6 & 85.3 & 72.5 \\
            FLAN-T5-base          & 87.0 & 25.7 & 86.7 & 87.8 & 78.5 & 80.3 & 40.4 & 81.1 & 81.1 & 68.5 & 67.2 & 14.6 & 69.0 & 66.4 & 56.7 & 81.9 & 42.9 & 82.2 & 82.9 & 71.8 \\
            T5-large              & 91.0 & \textbf{86.2} & 87.7 & \textbf{90.5} & 80.6 & 85.1 & \textbf{80.7} & 82.4 & 84.7 & 71.0 & 72.5 & \textbf{71.9} & \textbf{70.7} & \textbf{71.8} & 59.6 & 86.4 & \textbf{81.7} & 83.3 & 86.1 & 74.1 \\
            FLAN-T5-large         & \textbf{91.2} & 85.1 & \textbf{87.9} & 90.4 & \textbf{80.7} & \textbf{86.1} & 79.8 & \textbf{83.1} & \textbf{84.8} & \textbf{72.3} & \textbf{72.9} & 71.2 & 70.5 & 71.4 & \textbf{61.0} & \textbf{87.2} & 80.6 & \textbf{83.8} & \textbf{86.2} & \textbf{75.0} \\
            GPT-3.5               & 80.3 & 78.8 & 78.8 & 80.6 & 73.3 & 70.8 & 70.2 & 70.7 & 71.4 & 64.2 & 63.1 & 63.9 & 62.1 & 61.7 & 50.9 & 73.5 & 72.7 & 72.9 & 74.3 & 67.7 \\
            GPT-4                & 82.8 & 81.6 & 80.9 & 82.1 & 75.6 & 72.2 & 71.1 & 68.3 & 70.4 & 61.7 & 62.9 & 64.2 & 61.3 & 61.8 & 50.4 & 75.6 & 74.4 & 72.3 & 74.4 & 67.2 \\
            LLaMa-2-7B            & 34.3 & 29.6 & 37.6 & 36.5 & 39.2 & 28.6 & 24.6 & 31.1 & 30.3 & 29.1 & -18.2 & -25.8 & -14.6 & -15.0 & -13.3 & 30.8 & 27.1 & 34.1 & 32.9 & 35.1 \\
            Llemma-7B             & 75.7 & 73.9 & 73.0 & 74.9 & 62.6 & 63.6 & 63.8 & 61.9 & 63.4 & 52.6 & 59.7 & 63.4 & 59.3 & 58.9 & 49.9 & 67.3 & 66.7 & 65.6 & 67.2 & 56.2 \\
        \end{tabular}
    }
    \caption{Evaluation results with both in-distribution static scores (\textbf{S}) and those from the interventions (\textbf{VR}, \textbf{EE}, \textbf{AG}, \textbf{SR}).}
    \label{tab:combined_results}
\end{table*}

\renewcommand{\arraystretch}{1.1}
\begin{table*}[t]
    \centering
    \scalebox{0.6}{
            \begin{adjustbox}{margin=0pt 0pt 0pt 0pt}
        \begin{tabular}{@{}l|c|c|c|c|c|c|c|c|c|c|c|c|c|c|c|c|c|c|c|c|c|c|c|c|c|c|c|c|c|c@{}}
            & \multicolumn{5}{c|}{Syntax Errors} & \multicolumn{5}{c|}{Equality Errors} & \multicolumn{5}{c|}{Repeating Errors} & \multicolumn{5}{c|}{Redundant Errors} & \multicolumn{5}{c|}{Skipped steps} & \multicolumn{5}{c}{Verbose}\\
            \hline
            & \textbf{S} & \textbf{VR} & \textbf{EE} & \textbf{AG} & \textbf{SR} & \textbf{S} & \textbf{VR} & \textbf{EE} & \textbf{AG} & \textbf{SR} & \textbf{S} & \textbf{VR} & \textbf{EE} & \textbf{AG} & \textbf{SR} & \textbf{S} & \textbf{VR} & \textbf{EE} & \textbf{AG} & \textbf{SR} & \textbf{S} & \textbf{VR} & \textbf{EE} & \textbf{AG} & \textbf{SR} & \textbf{S} & \textbf{VR} & \textbf{EE} & \textbf{AG} & \textbf{SR}\\
            \hline
            T5-base & 20 & 32 & 7 & 19 & 25 & 0 & 3 & 2 & 2 & 0 & 7 & 36 & 35 & 8 & 2 & 1 & 15 & 18 & 1 & 8 & 69 & 46 & 44 & 65 & 155 & 7 & 73 & 28 & 4 & 0\\
            FLAN-T5-base & 2 & 43 & 4 & 3 & 1 & 1 & 23 & 0 & 0 & 0 & 7 & 33 & 24 & 5 & 3 & 2 & 13 & 25 & 0 & 10 & 95 & 196 & 64 & 92 & 168 & \textbf{5} & 33 & 22 & 4 & 0\\
            T5-Large & 11 & 11 & 14 & 9 & 15 & 0 & 3 & 8 & 0 & 0 & 5 & 28 & 33 & 8 & 3 & 1 & 1 & 21 & 2 & 4 & 53 & 41 & 41 & 53 & 140 & 9 & 40 & 33 & 13 & 2\\
            FLAN-T5-Large & \textbf{1} & 20 & 7 & 2 & 3 & \textbf{0} & 4 & 0 & 1 & 0 & 7 & 21 & 30 & 7 & 3 & 2 & 8 & 26 & 4 & 8 & \textbf{48} & 36 & 30 & 50 & 133 & 11 & 50 & 38 & 11 & 2\\
            GPT-3.5 & 4 & 0 & 1 & 1 & 3 & 2 & 2 & 2 & 3 & 2 & 3 & 4 & 3 & 3 & 1 & 0 & 2 & 1 & 0 & 1 & 96 & 91 & 81 & 98 & 128 & 51 & 37 & 41 & 26 & 28\\
            GPT-4 & \textbf{1} & 1 & 0 & 0 & 0 & \textbf{0} & 0 & 0 & 0 & 0 & \textbf{0} & 1 & 0 & 0 & 0 & \textbf{0} & 1 & 1 & 0 & 5 & 112 & 105 & 134 & 118 & 154 & 7 & 20 & 13 & 8 & 11\\
            LLaMa-2-7B & 6 & 3 & 5 & 6 & 9 & 4 & 8 & 7 & 6 & 17 & 14 & 11 & 28 & 9 & 35 & 6 & 2 & 3 & 1 & 8 & 99 & 91 & 115 & 107 & 219 & 21 & 17 & 33 & 15 & 27\\
            Llemma-7B & 13 & 11 & 13 & 11 & 35 & 17 & 12 & 13 & 9 & 26 & 94 & 139 & 153 & 108 & 243 & 1 & 11 & 3 & 1 & 16 & 114 & 110 & 103 & 134 & 110 & 92 & 125 & 120 & 92 & 181
        \end{tabular}
        \end{adjustbox}
        }
    \caption{Error counts for specific equation-level and derivation-level categories. \textbf{Syntax} refers to the number of equations with unbalanced brackets. \textbf{Equality} counts the number of equations without equality (or inequality) symbols. \textbf{Repeating} is the total number of repeated equations. \textbf{Redundant} is the number of equations where the LHS exactly matches the RHS. \textbf{Skipped steps} and \textbf{Verbose} respectively count the excess or reduced number of equations in the output compared to the reference derivation.}
    \label{tab:combined_error_counts}
\end{table*}

\section{Empirical Evaluation}

\paragraph{Empirical setup.} We evaluate a range of LLMs on the Derivation Generation task including T5 ~\cite{raffel2020exploring}, FLAN-T5~\cite{chung2024scaling}, GPT-3.5 and 4~\cite{brown2020language,achiam2023gpt}, LLaMa-2-7B~\cite{touvron2023LLaMa}, and Llemma-7B~\cite{azerbayev2023llemma}. We fine-tune base (220M) and large (770M) variants of T5 and FLAN-T5, while using the remaining models in a few-shot setting.
The evaluation occurs across 3 complementary methods: \textbf{(1.)} use of reference-based text generation metrics (Tab.~\ref{tab:combined_results}), \textbf{(2.)} error count as determined by searching model output for surface-level mathematical errors (Tab.~\ref{tab:combined_error_counts}), and \textbf{(3.)} a \textit{manual analysis} of models' reasoning accuracy across 750 total derivations (Tab.~\ref{tab:manual_scores}). Additional details on models, training, and metrics can be found online.\footnote{Further experimental details: \url{https://github.com/jmeadows17/deriving-equations-with-LLMs}}

\paragraph{Prompt design.}
We fine-tune and zero-shot prompt the T5 models following the template below, which corresponds to a ground truth sequence of equations: 

\vspace{10pt}
\begin{addmargin}[1em]{1em}
\textit{Given \;\; $q{(a)} = e^{a}$ \newline \newline and  \;\; $G{(a)} = - e^{a} + \frac{d}{d a} q{(a)}$, \newline \newline
then derive $- e^{a} + \frac{d}{d a} q{(a)} = 0$,\newline\newline
then obtain \;\; $e^{G{(a)}} = 1$}
\end{addmargin}
\vspace{10pt}

\noindent To few-shot prompt the decoder-only models (i.e., GPT, LLaMa and Llemma) we use the following design, where $n = 5$ is the number of in-context examples:

\noindent \rule{\linewidth}{0.3mm}
The following examples consist of a prompt (denoted by Prompt:) and a mathematical derivation (denoted by Derivation:). Each derivation contains LaTeX equations separated by "and".\newline

Prompt: [\texttt{Prompt} $1$]

Derivation: [\texttt{Derivation} $1$] \newline

$\vdots$\newline

Prompt: [\texttt{Prompt} $n$]

Derivation: [\texttt{Derivation} $n$] \newline

\noindent Now given the following prompt, generate the derivation. Ensure equations are split by the word "and". \newline

Prompt: $[\texttt{Evaluation Prompt}]$

\noindent \rule{\linewidth}{0.3mm}

\noindent This approach was chosen to minimise natural language in the generated output, and to force derivations into the desired format (LaTeX equations split by ``and''). Notably, only the \texttt{Evaluation Prompt} is perturbed, ensuring that the bulk differences in scores are not caused by changes to in-context examples, and evaluation is pair-wise consistent.

\subsection{Results with Text Generation Metrics}

Here we highlight the main results obtained using common text generation metrics including ROUGE~\cite{lin2004rouge}, BLEU~\cite{papineni2002bleu}, BLEURT~\cite{sellam2020bleurt}, and GLEU~\cite{mutton2007gleu}.

\paragraph{Small fine-tuned LMs outperform few-shot GPT-4 across all generation metrics.} On 2K examples (denoted by $(f)$) from the static set, FLAN-T5-large outperforms all models in all metrics. This minor advantage over T5-large may stem from further instruction fine-tuning (our prompt is an instruction). However this advantage over T5 does not extend to FLAN-T5-base, which scores lower than T5-base in all metrics. This may be due to fine-tuning instability observed in T5~\cite{asai-etal-2022-attempt}. We note that despite the success of the fine-tuned models, according to the metrics, we are not suggesting they are more suitable for equational reasoning than GPT, as the other sections of the evaluation reveal. 

The scores reported in Tab.~\ref{tab:combined_results} (without $(f)$) are evaluated on 100 examples from the static set \textit{explicitly containing integration and differentiation results in the prompt}, in order to fairly examine the effect of the Step Removal intervention which perturbs input by removing these results. The fine-tuned LMs score within 3 units of their previous scores, model rankings are preserved across all metrics, and we assume the GPT/LLaMa scores would report similarly for larger samples. With that said, according to all metrics, all fine-tuned models generally outperform all (5-shot) decoder models. Notably, vanilla LLaMa-2 scores are less than half of those obtained by Llemma, indicating the benefits of Llemma's fine-tuning on mathematical corpora~\cite{azerbayev2023llemma}. This difference is mirrored in the out-of-distribution scores. 


\subsection{Results with Template-based Error Detection}

In this section, we count the rate of different categories of errors by extracting equations from a model's derivation into an ordered list and adopting a set of metrics using both comparative and reference-free methods. 

We consider 6 error categories in total (Tab.~\ref{tab:combined_error_counts}) and sum the per-derivation error counts over all examples. 
For example, determining whether a model has either \textit{Skipped steps} or is too \textit{Verbose} (not necessarily ``errors'') occurs by comparing the length of the generated derivation with that of the ground truth reference. The number of \textit{Repeating equations} is determined by counting the number of uniquely generated equations in the output derivation (reference-free). The number of \textit{Equality} errors per derivation is calculated considering the number of equations that do not contain the ``='' token.


\paragraph{Fine-tuned FLAN-T5-Large and few-shot GPT-4 obtain the lowest rate of in-distribution errors.} Across the 6 error categories, both FLAN-T5-Large and GPT-4 possess the least \textbf{\textit{Syntax}} and \textbf{\textit{Equality}} errors. Results then diverge as FLAN-T5-Large repeats several equations \textbf{\textit{(Repeating)}} and outputs a couple of equations where the LHS is an exact string match with the RHS \textbf{\textit{(Redundant)}}, whereas GPT-4 \textbf{\textit{skips around twice as many steps}} as the largest fine-tuned models. This supports the claim that we are generally training models at a level of mathematical granularity \textit{below} that of the data used to train GPT-4.

\paragraph{Model rankings and perturbation difficulty based on template-based error count disagree with those determined by generation metrics.} At the top of the model rankings, on both the in-distribution and perturbed test sets, Tab.~\ref{tab:combined_error_counts} shows that GPT-4 and GPT-3.5 generally leapfrog the fine-tuned models in terms of low total error count, which disagrees with the scores obtained by the generation metrics. The LLaMa-based models remain at the bottom. 

In terms of perturbation difficulty, SR (Step Removal) is no longer the most challenging perturbation for fine-tuned models by error count, yet remains difficult for all other models. From this, \textit{FLAN-T5-Large does not outperform GPT-4 or GPT-3.5 on out-of-distribution examples}, but it generally \textit{does} outperform LLaMa-2 and Llemma. This suggests that for the fine-tuned models, certain interventions (such as SR) correspond to a clearly identifiable distribution of mathematical error types, such that the intervention may be inferred from the results. We discuss the implications of this after the manual evaluation.

\subsection{Human Evaluation of LLM Derivation Quality}

\begin{table}[t]
	\centering
	\scalebox{0.9}{
		\begin{tabular}{@{}l|c|c|c|c|c|c}
			& \textbf{S} & \textbf{VR} & \textbf{EE} & \textbf{AG} & \textbf{SR}\\
			\hline
GPT-4 & 98 & 96 & 92 & 100 & 80\\
FLAN-T5-Large & 98 & 68 & 62 & 92 & 76\\
Llemma-7B & 70 & 70 & 64 & 76 & 36
			\end{tabular}
		}
	\caption{Equational reasoning accuracy \% from human evaluation.} 
    \label{tab:manual_scores}
\end{table}

To recap, the generation metrics return scores and performance rankings via n-gram-based similarity measures between model derivations and ground truth references. In parallel, the error detection returns scores via a combination of reference-based and reference-free surface-level checks, that pick up on basic mathematical errors such as imbalanced brackets, missing equality signs, and repeating equations. What is lacking is an assessment of models' underlying operational reasoning that is largely independent of surface-level checks, and should be as close to a reference-free evaluation as possible within the scope of the task. We aim to provide such analysis in this section. Our approach to determining whether a model's derivation is coherent is as follows. A derivation:

\textbf{(1.)} must not include any equations that are inconsistent (and can not be made consistent by substituting a number for a variable, \textit{e.g.,} $\alpha = 1$).

\textbf{(2.)} must not include malformed equations with the exclusion of minor typos.

\textbf{(3.)} must include the exact goal equation, as this is always given in the prompt.

\textbf{(4.)} must apply operators correctly and in the correct order where operators are non-commutative.

\textbf{(5.)} may skip numerous steps, even premises, so long as a path may be reasonably derived between consistent equations.

\textbf{(6.)} may include irrelevant (but consistent) equations that do not contribute to the core path linking premises to the goal equation.

\textbf{(7.)} may repeat some equations.

\noindent The above marking scheme provides a very lenient framework which \textit{prioritises the consistency and operation-wise correctness of equations}. \newline \newline \noindent \textbf{Sampling derivations.} The results in Tab.~\ref{tab:manual_scores} were determined from a sample of 50 static derivations per model, with 4 corresponding perturbed derivations (VR, EE, AG, SR), totalling 250 per model. The 50 static derivations are sampled by ensuring the average ROUGE, BLEU, BLEURT, and GLEU scores over the sample aligns with that model's static score (\textbf{S}) in Tab.~\ref{tab:combined_results} to within 0.1 units. For instance, GPT-4's sample averages scores of 82.8, 72.2, 62.9, and 75.6, and the perturbed derivations are based on the static examples from this selection. 

We select derivations which are output by the best fine-tuned, GPT, and open-source models, totalling 750 examples. We convert equations into a easily readable format rendered in LaTeX, and either manually delete equations or fix any minor typos causing compilation errors. This document is available here.\footnote{\url{https://github.com/jmeadows17/deriving-equations-with-LLMs/blob/main/Derivation_Analysis.pdf}}

\paragraph{Fine-tuned FLAN-T5-Large and GPT-4 are matched in-distribution in terms of equation consistency and coherent use of operators.} Tab.~\ref{tab:manual_scores} illustrates how the fine-tuned and GPT models scored 98\% accuracy according to the lenient marking scheme. GPT's single incoherent derivation involved adding a variable to both sides of an equation, then later integrating and forcing that variable to be the constant of integration (breaking rule \textbf{(2.)}). FLAN-T5's incoherent derivation involved a sequence of equations which implied that $\frac{d\theta}{dq} = (\frac{d\theta}{dq})^q$, which is true only if $q = 1$ (breaking rule \textbf{(2.)}). 

\paragraph{Accounting for numerous surface-level error checks approximates human evaluation.} Despite the fact that one evaluation scheme focuses on equation consistency, while the other compares surface-level errors, the respective manual and template-based error results agree that, in-distribution, GPT-4 and FLAN-T5-Large are tied (with Llemma significantly underperforming). 
Regarding out-of-distribution evaluation, both manual and error-based scores agree that the {\textit{fine-tuned models are less capable than all generation metrics suggest}. In terms of intervention difficulty, both agree that SR \textit{(Step Removal)} is particularly challenging while AG \textit{(Alternative Goal)} is the least, and both schemes agree that the fine-tuned models are less affected by SR. This is not reflected by any of the generation metrics. 

Given this alignment between the manual evaluation (Tab.~\ref{tab:manual_scores}) and the template-based error detection (Tab.~\ref{tab:combined_error_counts}) spanning only 6 error types, and that more mathematically capable language models are less likely to hallucinate syntax errors and related trivially detectable artefacts, together this suggests that by accounting for a large $(>\!>6)$ number of surface-level errors, we can approximate human evaluation of LLM equational reasoning, \textit{at least} more faithfully than many canonical generation metrics. This can be achieved via the \textit{weighted average of counts over all surface-level categories}, where the category weights are empirically determined through comparison with rankings based on human evaluation.
\begin{figure}[t]
    \centering
    \includegraphics[width=1.0\linewidth]{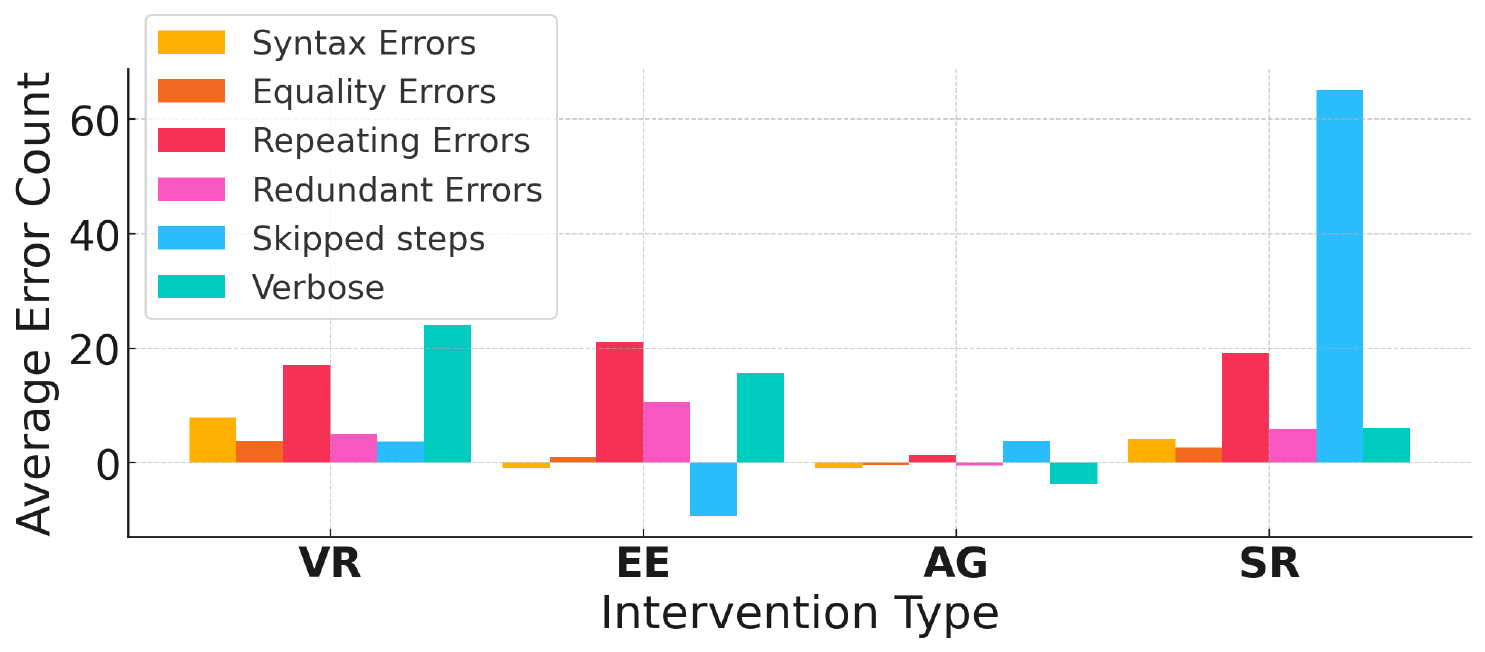}
    \caption{The average distributional footprint left by certain interventions.}
    \label{fig:footprints}
\end{figure}

\subsection{Controlling Equational Reasoning in LLMs} Supported by the manual and error-based evaluations, interventions are intrinsically linked to certain LLM hallucinations with varying degrees of association. 

We start defining such relationships by considering the rate of hallucination types $R(i, m, t)$ corresponding to an intervention $i$, error type $t$, model $m$, and the static rate $S(m, t)$ from Tab.~\ref{tab:combined_error_counts}. For instance, Fig.~\ref{fig:footprints} is characterised by:

\begin{equation}
    \delta(i,t) = \mathcal{N}\sum_{m}\big(R(i, m, t) -  S(m, t)\big)
\end{equation}

\noindent where $\mathcal{N}$ is a normalisation factor (reciprocal number of models considered). Hence $\delta(i,t)$ is the average error count for intervention $i \in \{\textbf{VR}, \textbf{EE}, \textbf{AG}, \textbf{SR}\}$ of type $t \in \{\text{Syntax}, \text{Equality}, ...\}$. If $\delta < 0$, then the intervention (on average) \textit{reduced} the rate of hallucinations of that type across the models, and vice versa. 

Furthermore, certain classes of interventions (\textit{e.g.,} injection of random noise, random token deletion) may depend on a practically continuous variable, $s$, that predictably varies the rate of certain errors. Hence, with some loss of information, intervention $i$ may be represented as a vector $\mathbf{x}_i(s) = \big(\mathbb{E}_t[\delta(i, t, s)], \, \sigma_t(\delta(i, t, s)), ...\big)$, where $\mathbb{E}_t[\delta(i, t, s)]$ and $\sigma_t(\delta(i, t))$ (etc.) are the expectation value and standard deviation of $\delta$ over the error types.

To find $s$ such that intervention $i$ likely improves the output quality over most hallucination types at that strength, we can write 
\begin{equation}
    \mathbb{E}_t[\delta(i, t, s)] + \varepsilon \sigma_t\big(\delta(i, t, s)\big) < 0
\end{equation}

\noindent where large $\varepsilon$ ensures that $\delta(i, t, s) < 0$ across a greater number of hallucination types $t$. 

The clear error distributions associated with each intervention in Fig.~\ref{fig:footprints} (characterised by Eq.~1) are averaged over all evaluated models, but most closely align with the fine-tuned T5 models. For this class of approaches, the interventions have a distinct effect on specific surface-level errors such as the rate of repeating equations or incorrect syntax. If these distributions $\mathbf{x}_i(s)$ may be further controlled by some variable $s$, we can define conditions for reducing surface-level error rates (\textit{e.g.,} Eq.~2), which correlates with improved derivation quality according to human evaluation.

\section{Related Work}

Our focus is evaluating and controlling the LLM-based~\cite{brown2020language,ahmed2022few,song2022llm,ge2023openagi,hu2023llm,yang2023harnessing,dubey2024LLaMa,meta2024introducing} generation of informal mathematical reasoning that resembles step-wise detailed equation derivations. While we focus on \textit{equation generation}, mathematical generation exists in various forms, and can be clustered into two main categories: approaches that consider formal languages, and those that consider informal mathematical natural language~\cite{meadows2023introduction,lu2022survey,zhong2022evaluating}. In the formal case, GPT-j~\cite{polu2020generative, polu2022formal}, LISA~\cite{jiang2021language}, and Baldur~\cite{first2023baldur} focus on Metamath and Isabelle/HOL proofs. For generation involving informal reasoning, an approach based on OpenAI's Codex~\cite{chen2021evaluating, drori2022neural} translates university-level problems into executable code, and generates solution explanations. Minerva~\cite{lewkowycz2022solving} is a PaLM~\cite{chowdhery2022palm} model trained on a large corpus of mathematical text, and solves university-level problems in applied math, outputting solutions in the form of mathematical natural language. NaturalProver~\cite{welleck2022naturalprover} generates similar solutions to proofs from a curated dataset~\cite{welleck2021naturalproofs}, and is most similar to our present work. However, our approach differs in a number of ways. Firstly, we focus exclusively on the generation of equational chains (in contrast to the inclusion of natural language statements). Our prompts and derivations are procedurally generated valid derivations in LaTeX, and many examples are guaranteed to include reasoning which is out-of-distribution with respect to other datasets, while containing up to 10 equations with wildcard symbols~\cite{zanibbi2016ntcir}. Lastly, our use of symbolic interventions follows from a previous approach~\cite{meadows2024symbolicframeworkevaluatingmathematical}, and we describe specific improvements in a later section.

\section{Conclusion}

To control and assess the fine-grained equation derivation capabilities of LLMs via prompt interventions, we first construct a dataset comprising 30k mathematically fine-grained prompt-derivation pairs using a symbolic data generation framework. We find fine-tuned models match or surpass the \textit{in-distribution} performance of few-shot GPT-4, despite a difference in parameter count of up to 3 orders of magnitude.

However, while all generation metrics suggest the fine-tuned models also outperform few-shot GPT-4 on perturbed data, the manual and template-based error detection methods both \textit{strongly disagree} with reference-based metrics, suggesting the latter lead to inappropriate representations of the relative effect of interventions and model capabilities. This strong alignment between human and error-based analysis suggests that extensive human evaluation can be approximated by accounting for numerous categories of surface-level errors.

We use these results to describe how a fundamental underlying mechanism relating distribution shifts to certain surface-level errors may be leveraged by varying the type and strength of prompt interventions to mitigate hallucination rates, which in turn facilitates control over the quality of LLM-based equational reasoning. 

Given a sufficiently large number of detectable mathematical hallucinations, an intervention with a variable strength that predictably controls the rate of certain errors, and an appropriate statistical condition, we can improve the quality of LLM reasoning post-training by experimentally determining an appropriate intervention strength.

\section*{Acknowledgements}
This work was partially funded by the Swiss National Science Foundation (SNSF) project NeuMath: Neural Discourse Inference over Mathematical Texts (200021\_204617).

\bibliography{aaai25}

\end{document}